# A New Clustering Method Based on Morphological Operations

Zhenzhou Wang*
College of Electrical and Electronic Engineering,
Shandong University of Technology, Zibo 255049, China
Email: (zzwangsia@yahoo.com)

*Abstract*—With the booming development of data science, many clustering methods have been proposed. All clustering methods have inherent merits and deficiencies. Therefore, they are only capable of clustering some specific types of data robustly. In addition, the accuracies of the clustering methods rely heavily on the characteristics of the data. In this paper, we propose a new clustering method based on the morphological operations. The morphological dilation is used to connect the data points based on their adjacency and form different connected domains. The iteration of the morphological dilation process stops when the number of connected domains equals the number of the clusters or when the maximum number of iteration is reached. The morphological dilation is then used to label the connected domains. The Euclidean distance between each data point and the points in each labeled connected domain is calculated. For each data point, there is a labeled connected domain that contains a point that yields the smallest Euclidean distance. The data point is assigned with the same labeling number as the labeled connected domain. We evaluate and compare the proposed method with state of the art clustering methods with different types of data. Experimental results show that the proposed method is more robust and generic for clustering two-dimensional or three-dimensional data.

## I. INTRODUCTION

CLUSTERING groups a set of data points into non-overlapping groups and it is essential in many machine learning and pattern recognition applications [1-8]. For instance, customer churn is managed by big data clustering in the service industry [1]. Daily load curves are clustered from smart-meter data in electricity data management [2]. K-means clustering is enhanced to determine the breaking points in traffic signal control [3]. Graph clustering is used in sensor array to locate sources [4]. Clustering based on fuzzy set theory [5-6], clustering based on spectral clustering [7] and clustering based on slope difference distribution [8] are used for image segmentation. The research of clustering methods has boomed in the past decades and so many clustering methods have been proposed to address different problems [9-13].

Although clustering is a general task to be solved, the algorithms might vary significantly and different clustering methods rely on completely different principles. Different principles have different merits and deficiencies. For instance, the density-based clustering method (GDBSCAN) [14] is robust in clustering data points with arbitrary shapes when the point densities within data classes could be distinguished from the point densities between data classes. Yet, GDBSCAN might fall into local partition within the same data class when point densities within classes vary greater than point densities between classes. K-means is a very popular clustering method and widely used in many fields [15-18]. The biggest merit of K-means is that it could be applied in high dimensional data clustering. Different algorithms have been proposed to improve the accuracy and efficiency of the K-means method. However, these proposed algorithms could not guarantee the convergence of the K-means to the optimal solution. The accuracy of K-means method is affected by the shape of the data points significantly. In addition, the K-means problem is an approximation problem without analytical solutions, which further limits its accuracy. Another problem of K-means is that when the dataset is extremely large, its iteration process might take too much time to converge. Enlightened by the K-means method, researchers have tried to propose other more effective iterative methods to address the specific clustering problems more robustly. For example, the authors in proposed a fuzzy clustering method that clustered the IRIS data more accurately than K-means method. The authors in proposed a simple and fast K-medoids clustering method that also achieved better accuracy in clustering the IRIS data. In addition, there are many other popular clustering methods that have been proposed in the past decades, e.g. Fuzzy c-means clustering method, mean-shift clustering method and spectral clustering method. All these clustering methods have inherent advantages and disadvantages. In this paper, we will give a comparative study of these state of the art methods with some generic data points. In addition, we will propose a new clustering method based on the adjacency of data points.

By nature, the adjacent data points in the same class should be closer to each other than the adjacent data points between different classes. Therefore, we propose to cluster the data points based on the adjacency of the data points. Firstly, all data points are shifted and scaled in a positive range. The positive data points are rounded to be integers and then transformed into a grid. The location of each rounding data point in the grid is assigned with the value one and all the other locations are assigned with the value zero. The locations assigned with the value one are expanded by assigning all its adjacent locations with the value of one until the locations of the data points in the same class merge into one connected domain. All the connected domains are then labeled with sequential numbers, e.g. one, two, three and so on. For each positive data point, its smallest Euclidean distance to each labeled connected domain is calculated. The positive data point is labeled with the same number as its nearest connected domain. After all the positive data points are labeled with a number, they are transformed back into the original range.

## II. THE PROPOSED METHOD

For a given two-dimensional or three-dimensional data with the length of $N$, $p = (p_1(i), p_2(i)), i = 1, 2, ..., N$ or $p = (p_1(i), p_2(i), p_3(i)), i = 1, 2, ..., N$, all the data points are shifted and scaled by the following equation to make sure that all of them are positive.

$$P_i = R \times (p_i - \min(p_i)) / \max(p_i); i = 1, 2 \text{ or } 1, 2, 3 \quad (1)$$

where $R$ is the range of the scaled data points. $P_i$ denotes the shifted and scaled data set. If the data set is two-dimensional, the rounding data set is computed as:

$$[P] = ([P_1(i)], [P_2(i)]), i = 1, 2, ..., N \quad (2)$$

If the data set is three-dimensional, the rounding data set is computed as:

$$[P] = ([P_1(i)], [P_2(i)], [P_3(i)]), i = 1, 2, ..., N \quad (3)$$

The indexes in the first dimension of the two-dimensional or three-dimensional dataset are denoted by the set $X = \{1, 2, ..., R\}$. The indexes in the second dimension of the two-dimensional or three-dimensional dataset are denoted by the set $Y = \{1, 2, ..., R\}$. The indexes in the second dimension of the three-dimensional dataset are denoted by the set $Z = \{1, 2, ..., R\}$. If the dataset is two-dimensional, a two-dimensional grid is initialized and updated by the following equation.

$$G(x, y) = 0, \forall x \in X, y \in Y \quad (4)$$
$$G(x, y) = 1, \forall (x, y) \in [P] \quad (5)$$

If the dataset is three-dimensional, a three-dimensional grid is initialized and updated by the following equations.

$$G(x, y, z) = 0, \forall x \in X, y \in Y, z \in Z \quad (6)$$
$$G(x, y, z) = 1, \forall (x, y, z) \in [P] \quad (7)$$

For each location with the value of 1, its adjacent locations are also assigned with the value of 1 by the following equation.

$$G' = \{z \mid (\bar{B})_z \cap G \neq \emptyset\} \quad (8)$$

where $B$ is the structuring element and $\bar{B}$ is its reflection. The structuring element is a unit disk if the data set is two-dimensional and is a unit sphere if the data set is three-dimensional.

Eq. (8) is repeated to connect the adjacent data points in the same class and generate a connected domain. To avoid the situation in which data points belonging to the same class are divided into different small connected domains, a domain filter is designed to remove small and invalid domains.

Immediately following the adjacent points connecting process by Eq. (8), all the domains in the grid are labeled to generate a labeled grid $G_L$. The coordinate set $(X_j, Y_j), j = 1, 2, ..., M$ of the labeled domain for the two-dimensional data set is computed as:

$$(X_j, Y_j) = \{(x, y) \mid x \in X, y \in Y, G_L(x, y) = j\} \quad (9)$$

The coordinate set $(X_j, Y_j, Z_j), j = 1, 2, ..., M$ of the labeled domain for the three-dimensional data set is computed as:

$$(X_j, Y_j, Z_j) = \{(x, y, z) \mid x \in X, y \in Y, y \in Y, G_L(x, y, z) = j\} \quad (10)$$

The size threshold to separate the small domains and the large domains is calculated as:

$$T_S = \frac{1}{M}\sum_{j=1}^{M}|X_j| = \frac{1}{M}\sum_{j=1}^{M}|Y_j| = \frac{1}{M}\sum_{j=1}^{M}|Z_j| \quad (11)$$

Where $|X_j|, |Y_j|$ or $|Z_j|$ denotes the total number of the element contained in the set $X_j, Y_j$ or $Z_j$, i.e. the total number of pixels in the $j$th labeled domain.

The coordinates of the small and invalid domains for the two-dimensional data set are computed as:

$$(X_S, Y_S) = \bigcup_{j=1}^{M} \{(x, y) \mid x \in X, y \in Y, |X_j| < T_S\} \quad (12)$$

The coordinates of the small and invalid domains for the three-dimensional data set are computed as:

$$(X_S, Y_S, Z_S) = \bigcup_{j=1}^{M} \{(x, y, z) \mid x \in X, y \in Y, z \in Z, |X_j| < T_S\} \quad (13)$$

The grid is updated by the following equations to remove the small and invalid domains in two-dimension or three-dimension respectively.

$$G'(x, y) = 0, \forall (x, y) \in (X_S, Y_S) \quad (14)$$
$$G'(x, y, z) = 0, \forall (x, y, z) \in (X_S, Y_S, Z_S) \quad (15)$$

After small and invalid domains are removed, the remaining large domains are labeled again. If the total number of the labeled connected domain is equal to the inputted number of the clusters, the domain connecting process stops. If not, the above process (Eqs. 8-15) is repeated until the number of the labeled connected domains is equal to the inputted number of the clusters.

The points in the labeled connected domains are denoted as $C(l) = (C_1(l, k), C_2(l, k)), k = 1, 2, ..., K_l, l = 1, ..., L$ for the two-dimensional data set and denoted as $C(l) = (C_1(l, k), C_2(l, k), C_3(l, k)), k = 1, 2, ..., K_l, l = 1, ..., L$ for three-dimensional data set. $L$ denotes the total number of labeled connected domains and $K_l$ denotes the total number of point in the $l$th connected domain. For the $i$th scaled point in the two-dimensional domain, its Euclidean distance to the $k$th point in the $l$th labeled connected domain is calculated as:

$$d(l) = \left((P_1(i) - C_1(l, k))^2 + (P_2(i) - C_2(l, k))^2\right)^{1/2} \quad (16)$$

For the $i$th scaled point in the three-dimensional domain, its Euclidean distance to the $k$th point in the $l$th labeled connected domain is calculated as:

$$d(l) = \left( \begin{array}{l} (P_1(i) - C_1(l,k))^2 + (P_2(i) - C_2(l,k))^2 \\ + (P_3(i) - C_3(l,k))^2 \end{array} \right)^{1/2} \quad (17)$$

For the $i$th scaled point, its nearest Euclidean distance to the $l$th labeled connected domain is calculated by the following equations.

$$\bar{k} = \arg\min_{k=1,2,\ldots,K_l} |d(l)| \quad (18)$$

$$d_l = d(\bar{k}) \quad (19)$$

The label of the $i$th scaled point is calculated as:

$$\bar{l} = \arg\min_{l=1,2,\ldots,L} |d_l| \quad (20)$$

There are also situations in which the data set contains noise. In such cases, a distance threshold $T_d$ is used to distinguish the data from the noise. The label of the $i$th scaled point is calculated as:

$$\bar{l} = \begin{cases} \arg\min_{l=1,2,\ldots,L} |d_l|, & if\ |d_l| < T_d \\ 0, & else \end{cases} \quad (21)$$

The value of $T_d$ could be specified based on the range of the data manually or calculated by the follow Equation.

$$T_d = \frac{1}{N} \sum_{i=1}^{M} \left( d_l(i) - \bar{d_l} \right)^2 \quad (22)$$

$$\bar{d_l} = \frac{1}{N} \sum_{i=1}^{M} d_l(i) \quad (23)$$

where $M$ is the total number of minimum distances.

After all the scaled points $P_i; i=1,2\ or\ 1,2,3$ are labeled as $P_i^l; i=1,2\ or\ 1,2,3; l=1,\ldots,L$, they are transformed back to their original range by the following equation:

$$p_i' = P_i^l \times \max(p_i)/R + \min(p_i); i=1,2\ or\ 1,2,3 \quad (24)$$

where $p' = (p_1'(i), p_2'(i), p_3'(i)), i=1,2,\ldots,N$ or $p' = (p_1'(i), p_2'(i)), i=1,2,\ldots,N$ are the clustered data points.

## III. EXPERIMENTAL RESULTS

### A. Comparison with State of the Art Methods

In this section, we will compare the proposed clustering method with state-of-the-art clustering methods both qualitatively and quantitatively. The compared clustering methods include the density-based clustering method (GDBSCAN), K-means method, the K-medoid method, the Ward based agglomerative clustering method, the MinKowSKi based agglomerative method, the mean-shift method and the spectral clustering method. The first tested dataset is synthesized by MATLAB as follows. Two hundred and fifty random points with the variance of 0.6 are generated around the center (1,1), (-1,-1) and (1,-1) respectively. The process of points connecting by the proposed method is shown in Figure 1 (a) and (b).

The final clustering results generated by the proposed method are shown in Figure 3 (a). The numbers of clustered points in three classes are 250, 246 and 254 respectively. The clustering results of the density-based clustering method (GDBSCAN) are shown in Figure 3 (b). The numbers of clustered points in three classes are 6, 462 and 217 respectively. In addition, 65 points are clustered as noise. The clustering results by the K-means method are shown in Figure 3 (c). The numbers of clustered points in three classes are 246, 252 and 252 respectively. The clustering results by the K-medoid method are shown in Figure 3 (d). The numbers of clustered points in three classes are 245, 251 and 254 respectively. The clustering results by the mean-shift method are shown in Figure 3 (e). The numbers of clustered points in three classes are 225, 258 and 267 respectively. The clustering results by the spectral clustering method are shown in Figure 3 (f). The numbers of clustered points in three classes are 251, 246 and 253 respectively. The clustering results by agglomerative clustering (Ward) and agglomerative clustering (MinKowSKi) are shown in Figure 3 (g) and (h) respectively. As can be seen, only the Ward method works correctly. The numbers of clustered points in three classes by Ward method are 256, 246 and 248 respectively. From the qualitative results, the density-based and MinKowSKi based clustering methods could not cluster the synthesize data correctly because they fell into local partition within the same data class. For those that could cluster the synthesized data correctly, only the proposed method clustered one class of data points with the hundred percent accuracy of 250/250. For better comparison, the clustering accuracies of different methods for this data set are shown in Table 1. Where GT denotes the ground truth and the accuracy is measured by the true positive (TP) rate. As can be seen, the proposed method and spectral clustering method achieved the best accuracy while the proposed method achieved second-best accuracy.

Table 1. Quantitative comparison of the accuracies of the clustering methods for the first synthesized data.

| GT | 250 | 250 | 250 | Accuracy |
|---|---|---|---|---|
| Proposed | 250 | 246 | 254 | **0.9893** |
| DBSCAN | 6 | 462 | 217 | 0.2107 |
| K-means | 246 | 252 | 252 | 0.9893 |
| K-medoid | 245 | 251 | 254 | 0.9867 |
| Mean-shift | 225 | 258 | 267 | 0.9333 |
| Spectral | 251 | 253 | 246 | 0.9893 |
| Ward | 256 | 246 | 248 | 0.984 |
| MinKowSKi | 1 | 1 | 748 | 0.0027 |

Table 2. Quantitative comparison of the accuracies of the clustering methods for the second synthesized data.

| GT | 200 | 200 | 600 | Accuracy |
|---|---|---|---|---|
| Proposed | 200 | 208 | 592 | **0.984** |
| DBSCAN | 198 | 13 | 786 | 0.625 |
| K-means | 200 | 221 | 579 | 0.958 |
| K-medoid | 200 | 221 | 579 | 0.958 |
| Mean-shift | 200 | 220 | 580 | 0.96 |
| Spectral | 200 | 321 | 479 | 0.758 |
| Ward | 200 | 212 | 588 | 0.976 |
| MinKowSKi | 199 | 1 | 800 | 0.199 |

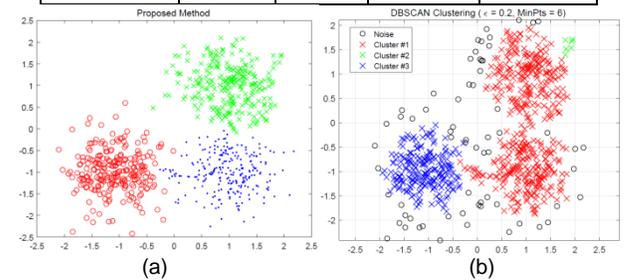

(a)     (b)

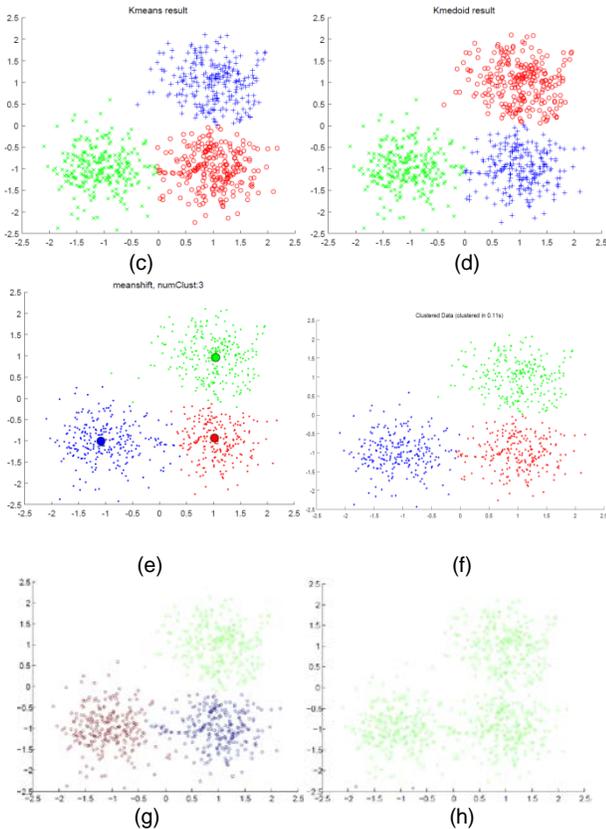

**Figure 3:** Comparison of the clustering results by different methods with the first synthesized dataset. (a) Result of the proposed; (b) Result of DBSCAN; (c) Result of K-means; (d) Result of K-medoid; (e) Result of Mean-shift; (f) Result of Spectral clustering; (g) Result of Agglomerative clustering (Ward); (h) Result of Agglomerative clustering (MinKowSKi).

We use another synthesized dataset downloaded from the internet to compare the proposed method with state of the art methods. The computed accuracies by different methods are shown in Table 2. As can be seen, accuracy of the proposed method is significantly more accurate than state of the art methods. We also show the qualitative results in Figure 4.

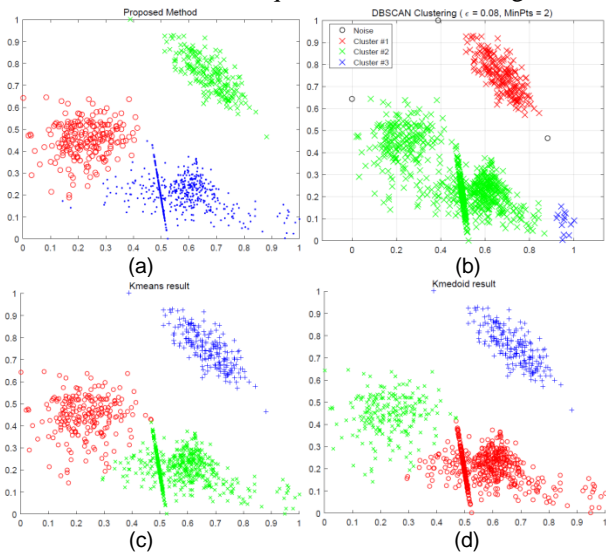

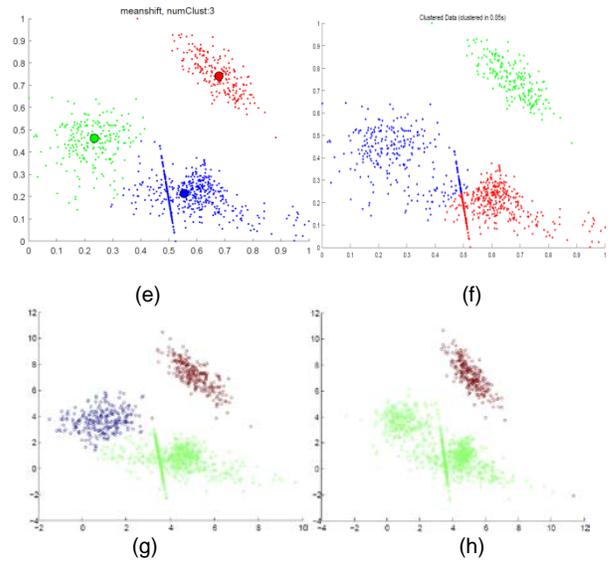

**Figure 4:** Comparison of the clustering results by different methods with another synthesized dataset. (a) Result of the proposed; (b) Result of DBSCAN; (c) Result of K-means; (d) Result of K-medoid; (e) Result of Mean-shift; (f) Result of Spectral clustering; (g) Result of Agglomerative clustering (Ward); (h) Result of Agglomerative clustering (MinKowSKi).

## IV. CONCLUSION

In this paper, a robust method is proposed to cluster both two-dimensional and three-dimensional data points. The proposed method use morphological operations to connect adjacent points to form connected domains and the data points are then clustered based on their distances to the labeled connected domains. The proposed method is compared with state of the art methods with different types of datasets. Experimental results show that the proposed method could cluster more types of two-dimensional or three-dimensional datasets correctly than state of the art methods. Among all the clustering methods that could cluster a specific type of dataset correctly, the proposed method achieved similar or significantly better accuracy.